\pdfoutput=1

\documentclass[11pt]{article}

\usepackage[preprint]{acl}

\usepackage{times}
\usepackage{latexsym}
\usepackage{tabularx} 
\usepackage{adjustbox}
\usepackage[T1]{fontenc}

\usepackage[utf8]{inputenc}

\usepackage{microtype}
\usepackage{longtable}
\usepackage{inconsolata}

\usepackage{makecell}
\usepackage{graphicx}
\usepackage{array}
\usepackage{booktabs}
\usepackage{relsize}
\usepackage{fvextra}
\usepackage{tcolorbox}
\tcbuselibrary{breakable}
\usepackage{listings}
\graphicspath{{images/}}
\title{SPeCtrum: A Grounded Framework for Multidimensional Identity Representation in LLM-Based Agent}

\author{
 \textbf{Keyeun Lee\textsuperscript{1 2}},
 \textbf{Seo Hyeong Kim\textsuperscript{1 2}},
 \textbf{Seolhee Lee\textsuperscript{1 2}},
 \textbf{Jinsu Eun\textsuperscript{1}},
\\
 \textbf{Yena Ko\textsuperscript{2}},
 \textbf{Hayeon Jeon\textsuperscript{1}},
 \textbf{Esther Hehsun Kim\textsuperscript{1 2}},
 \textbf{Seonghye Cho\textsuperscript{1}},
\\
 \textbf{Soeun Yang\textsuperscript{2}},
 \textbf{Eun-mee Kim\thanks{Corresponding authors}\textsuperscript{2}},
 \textbf{Hajin Lim\footnotemark[1]\textsuperscript{1 2}}
\\
 \textsuperscript{1}hci+d Lab, \textsuperscript{2}Department of Communication \\
 Seoul National University\\
    \texttt{\{kieunp, eunmee, hajin\}@snu.ac.kr}
}

\begin{document}
\maketitle

\begin{abstract}
Existing methods for simulating individual identities often oversimplify human complexity, which may lead to incomplete or flattened representations. To address this, we introduce \textbf{SPeCtrum}\footnote{Code and data available at: \url{https://github.com/keyeun/spectrum-framework-llm}}, a grounded framework for constructing authentic LLM agent personas by incorporating an individual's multidimensional self-concept. SPeCtrum integrates three core components: \textbf{Social Identity (S), Personal Identity (P), and Personal Life Context (C)}, each contributing distinct yet interconnected aspects of identity. To evaluate SPeCtrum’s effectiveness in identity representation, we conducted automated and human evaluations. Automated evaluations using popular drama characters showed that Personal Life Context (C)—derived from short essays on preferences and daily routines—modeled characters' identities more effectively than Social Identity (S) and Personal Identity (P) alone and performed comparably to the full SPC combination. In contrast, human evaluations involving real-world individuals found that the full SPC combination provided a more comprehensive self-concept representation than C alone. Our findings suggest that while C alone may suffice for basic identity simulation, integrating S, P, and C enhances the authenticity and accuracy of real-world identity representation. Overall, SPeCtrum offers a structured approach for simulating individuals in LLM agents, enabling more personalized human-AI interactions and improving the realism of simulation-based behavioral studies.

\end{abstract}
\section{Introduction}
\begin{quote}
\textit{Every man is more than just himself; he also represents the unique [..] point at which the world's phenomena intersect. - Herman Hesse (1919)}
\end{quote}
Identity is a complex, multifaceted concept that encompasses an individual's social, personal, and contextual attributes \citep{hall2015cultural,hesse2013demian, mead1934mind}. Recent advances in large language models (LLMs) have inspired their use in simulating individual behavior and thought processes \citep{park_social_2022, park_generative_2023, kang_values_2023}. However, some methods for simulating identities result in LLM agents that portray stereotypical characteristics of certain demographic groups \citep{argyle_out_2023,gupta_bias_2023, demollm2024}, often oversimplifying the complexity and diversity of individual human beings \citep{cheng_compost_2023, santurkar_whose_2023, petrov_limited_2024, bommasani2022}.

\begin{table}[t]
\centering
\fontsize{8.5pt}{10pt}\selectfont
\begin{tabular}{>{\raggedright\arraybackslash}m{1.6cm} >{\raggedright\arraybackslash}m{2.8cm} >{\raggedright\arraybackslash}m{2cm}}
\toprule
Component & Description & Source \\ \midrule
Social Identity (S) & One's innate and acquired qualities linked to a social group & 19 Demographic questionnaire items \\ \midrule
Personal Identity (P) & One's psychological traits and values & BFI-2-S and PVQ scale \\ \midrule
Personal Life Context (C) & One's unique realization of identity & Short essays (preference, daily routines) \\ \bottomrule
\end{tabular}
\caption{Components of the SPeCtrum Framework.}
\label{tab:components}
\end{table}

\begin{figure*}[h]
    \centering
    \includegraphics[width=\textwidth]{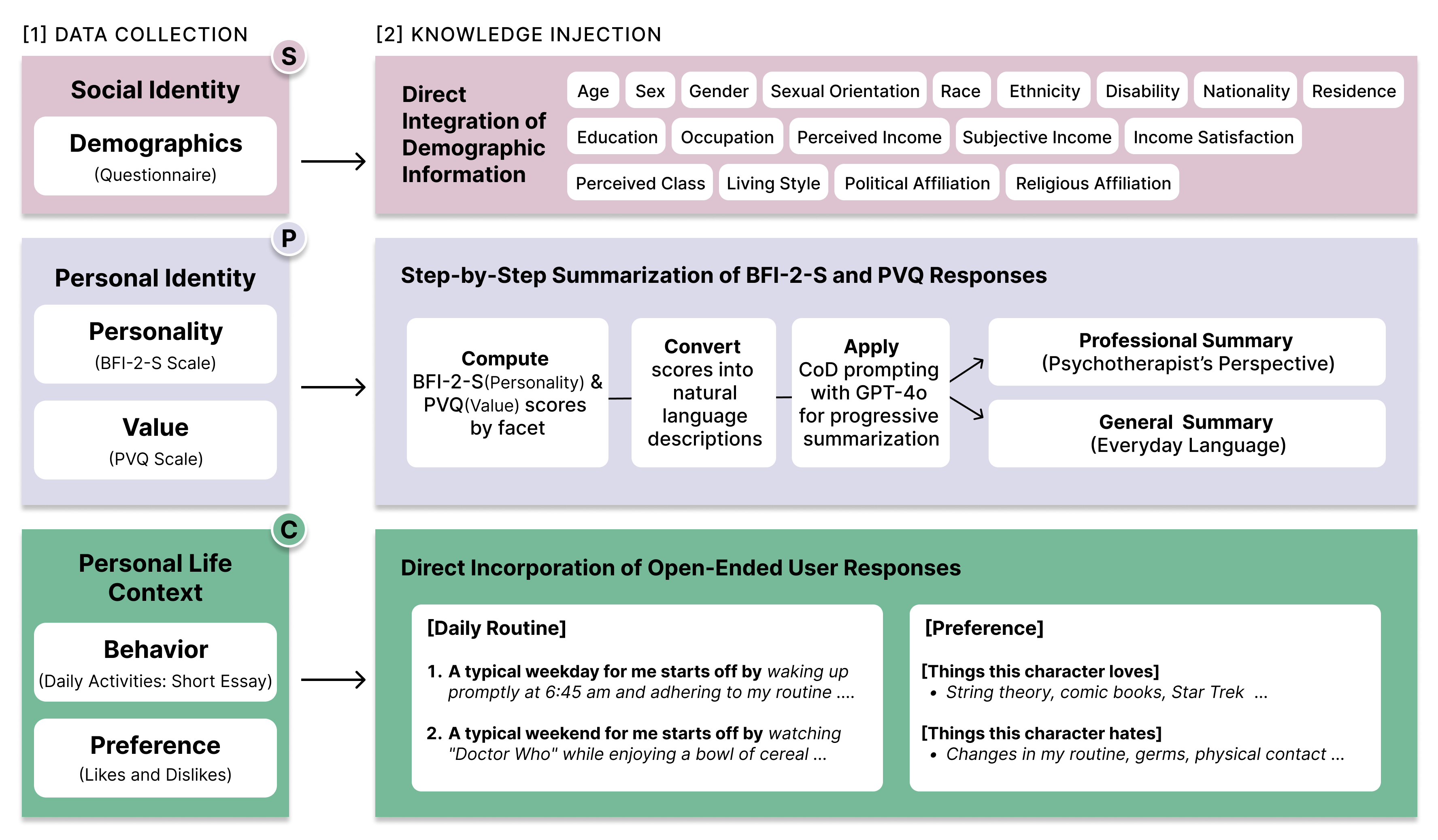}
    \caption{Overview of the SPeCtrum Framework for Multidimensional Identity Representation}
    \label{fig:1}
\end{figure*}

To address this limitation, we introduce the \textbf{SPeCtrum} framework (\textbf{S}ocial Identity, \textbf{P}ersonal Identity, and Personal Life \textbf{C}ontext), a grounded approach for developing LLM agents that effectively reflect the multidimensional nature of real-world individuals (see Table \ref{tab:components}). Grounded in social science approaches to one's self-concept \citep{Jones2000ACM, mead1934mind}, the SPeCtrum framework constructs identity through three key components: \textbf{Social Identity} (S), which refers to one's innate and acquired qualities linked to a social group, captured through demographic questionnaires; \textbf{Personal Identity} (P), which encompasses one's psychological traits and values,  assessed using established scales; and \textbf{Personal Life Context (C)}, representing one’s unique realization of identity, gathered through short open-ended essays on daily routines and personal preferences. 

To validate the SPeCtrum framework in representing one's self-concepts, we conducted both automated and human evaluations across different combinations of S, P, and C. For the automated evaluation, we created a dataset of popular drama characters and used the ``Guess Who test'' for character identification, along with the ``Twenty Statement Test'' (TST) using four LLMs: GPT-4o, GPT-3.5 Turbo, Claude-3.5-Sonnet, and Claude-3-Sonnet. In the human evaluations, 80 participants compared four types of agents—built on S, P, C, and SPC—based on how well each reflected their own self-perceptions.

The automated evaluation showed that Personal Life Context (C) was the most effective in capturing character identity, outperforming S and P, and performing comparably to the SPC combination. A follow-up experiment further tested whether C alone could infer S and P aspects. Results indicated that C content alone could reasonably infer demographic information (S) and personality traits and values (P) of drama characters.

Building on these findings, we conducted a human evaluation to assess whether this pattern held for real-world individuals who would be less prominently represented in LLMs' training data. The results indicated that while C continued to outperform S and P, the SPC combination provided a more comprehensive representation of self-concept for real-world individuals than C alone. Inference tests that derived S and P from C showed lower overall accuracy compared to inferences made from drama characters.

Such divergence between automated and human evaluations highlights the complexity of representing human identity, emphasizing the need for the full combination of S, P, and C as a comprehensive approach to capture self-concepts in real-world individuals accurately. Taken together, our evaluations validate the SPeCtrum framework as an effective foundation for representing multidimensional self-concepts and highlight its potential to enhance human-AI interactions and social simulations through more personalized and authentic identity representation.

Our primary contributions are as follows:
\begin{itemize}
    \item Introduce the SPeCtrum framework to facilitate the authentic and structured representation of real-world individuals in LLM agents.
    \item Demonstrate the effectiveness and potential of SPeCtrum through systematic evaluations, including automated evaluations using popular drama characters and human evaluations involving real-world individuals.
\end{itemize}

\section{Related Work}

Recent advancements in LLMs have significantly expanded their application across diverse academic fields such as social science \citep{aher_using_2023, gao_s3_2023}, behavioral economics \citep{horton_large_2023}, and human-computer interaction \citep{hamalainen_evaluating_2023}, primarily through the creation and deployment of diverse agent personas. These LLM-created personas aim to simulate complex human and social behaviors \citep{park_social_2022, park_generative_2023}, enabling the development of increasingly personalized applications such as recommendation systems \citep{wang_user_2023}.

Existing frameworks for persona creation primarily emphasize isolated human traits, focusing mainly on socio-demographic characteristics \citep{chen_empathy_2024, zhang_speechagents_2024, chuang-etal-2024-simulating} for modeling specific human subpopulations \citep{argyle_out_2023}. Although researchers have expanded these frameworks to incorporate other dimensions such as personality traits \citep{jiang-etal-2024-personallm, liu2024skepticism, xie_human_2024, yuan_evaluating_2024} and value systems \citep{zhou_sotopia_2024, xie2024largelanguagemodelagents, kang_values_2023}, their approaches often yield biased or incomplete representations, as evidenced by homogeneous depictions of socially underrepresented groups \citep{petrov_limited_2024, gupta_bias_2023, deusex_2024, cheng_compost_2023, lee_large_2024}.

Research in social psychology demonstrates that an individual's self-concept emerges from the dynamic interplay of multiple identity dimensions, including personal traits, social interactions, and lived experiences \citep{mead1934mind}. This fundamental understanding highlights the critical importance of incorporating such multidimensional aspects in developing LLM agents that authentically reflect real-world individuals and their behavioral and thought patterns \citep{xiao_how_2023}.

To address these limitations, we introduce the SPeCtrum framework (see Figure \ref{fig:1}), which enables structured and authentic persona representations through (a) identifying essential elements of multidimensional self-concept based on social science theories and research methodologies and (b) developing systematic pipelines for integrating diverse identity sources. Moving beyond the dominant focus on isolated traits, our framework emphasizes the dynamic interactions between identity components to create LLM-based agents that capture the rich complexity of real-world individuals.

\section{SPeCtrum: Framework for Multidimensional Identity Representation in LLM-based Agents}

The SPeCtrum framework is grounded in the concept of \textbf{self-concept}, which is an individual's set of beliefs and perceptions about themselves. This encompasses their beliefs, values, abilities, and attributes \citep{bracken1996handbook,markus1987dynamic,oyserman_self-concept_2001}. In particular, researchers have posited that self-concept primarily comprises two core components: Social Identity and Personal Identity \citep{Jones2000ACM, nario-redmond_social_2004}.

\paragraph{Social Identity (S)}
Social Identity refers to the shared characteristics of an individual as a member of various groups, including innate categories (e.g., gender, race) and acquired traits (e.g., education, occupation) \cite{oyserman_self-concept_2001}. These factors shape one’s self-concept and social interactions. To model this in LLM agents, we compiled a \textbf{set of 19 questions} focused on demographics and socio-economic status. This data is incorporated as a key element in constructing the social aspect of an individual's self-concept (see \ref{tab:demo}).

\paragraph{Personal Identity (P)}
Personal Identity encompasses deeper psychological traits and values, representing qualities individuals often consider core to their ``inner self'' \citep{Jones2000ACM, fearon1999identity}. It includes personal attributes and characteristics such as personality and value systems \citep{fearon1999identity}.

\textbf{\textit{Personality}} is defined as the dynamic organization of psychophysical systems within an individual that determines their adaptation to the environment \citep{allport1937personality}. By incorporating the personality factor, we aim to capture the primary personality traits that shape an individual's unique thought patterns, emotions, and behavior \citep{corr&matthews2009, weinberg_foundations_2019}. To model this in LLM-based agents, we used the 30-item Big Five Inventory-2-Short Form (\textbf{BFI-2-S}) \citep{bfi_short_2017}, a widely used and well-validated measure of the personality traits (e.g., extraversion).

\textbf{\textit{Values}} play a crucial role in shaping one's identity, influencing not only moment-to-moment behaviors but also guiding overarching life orientations \citep{schwartz1994there}. To integrate an individual’s value system into LLM agents, we employed the 21-item Portrait Values Questionnaire (\textbf{PVQ}) \citep{schwartzpvq2009basic}, which evaluates values across ten dimensions (e.g., hedonism, achievement, power). Incorporating the PVQ results provides a comprehensive perspective on an individual's values and their role in shaping personal identity.

\paragraph{Personal Life Context (C)}
Lastly, we incorporated Personal Life Context (C) to provide a more nuanced and dynamic understanding of how social and personal identity are expressed and enacted in an individual's daily life \cite{chen_meta-analysis_2024, frederickx2014role, schwartz1994there}.
To integrate Personal Life Context into our framework, we elicited \textbf{two short open-ended questions}: 1) one's \textbf{preferences} (listing five things one loves and hates) and 2) a short essay detailing \textbf{typical daily routines}, using the Behavioral Essay format \citep{boyd_values_2021}, which asks respondents to describe their typical activities on weekdays and weekends (see \ref{tab:openended}).
We expected that personal preferences would provide insight into one's tastes and interests while daily routines would reveal time management and priorities, thereby enhancing the realism and authenticity of LLM agents.

\subsection{Knowledge Injection Process}
\label{sec:knowledge_injection_pipieline}

We integrated the aforementioned information sources to elicit S, P, and C aspects. In injecting this data into LLMs, S component data, such as gender and race, were formatted into a list based on structured demographic questionnaires (see \ref{sec:appendix_profile_example_S_A1}). For C, which was collected in an open-ended format, we integrated the data directly into prompts without further processing (see \ref{sec:appendix_profile_example_C_A3}). This approach leverages evidence that contextual details, such as character utterances and writing styles, help produce a more nuanced, authentic representation \citep{han_meet_2022, ahn_mpchat_2023, shao_character-llm_2023}.

In processing the scores from the BFI-2-S and PVQ on a 1-7-point Likert scale into the P component, we followed the methodology of \citet{serapio-garcia_personality_2023} to avoid summaries that merely echo the verbatim of the scale items (see Figure \ref{fig:1}). 

To achieve this, we first averaged facet scores for each scale and converted them into natural language descriptions. For example, an Extraversion score of 3 was phrased as ``\textit{Extraversion is slightly below average.}'' Next, we applied the Chain of Density (CoD) technique \citep{adams_sparse_2023} with GPT-4o, the state-of-art model (SOTA) at the time, for progressive summarization. This involved first generating a technical summary of the facet descriptions and then iteratively condensing it into denser and more insightful summaries. Through this process, the final personality and value description produced two overviews of an individual's personality and value system: one in expert terminology (psychotherapist's language) and the other in everyday language (see \ref{sec:appendix_profile_example_P_A2}), providing a comprehensive understanding of one's personal identity.

Through this process, we constructed an individual’s holistic profile encompassing S, P, and C. We then prompted LLMs to ``embody the character,'' focusing on how a character’s traits manifest in both personal and social contexts without directly referencing the provided data (see \ref{sec:appendix_profile_example_base_prompt}).

\section{Automated Evaluation with Fictional Characters}
To examine the viability of the framework in capturing and representing an individual’s self-concept, we first conducted an automated evaluation using popular U.S. drama characters. Specifically, we aimed to assess how well each component (S, P, and C), both individually and in combination, represents key aspects of an individual’s identity.

In doing so, we performed a comprehensive analysis to evaluate the contribution of each component by testing all possible combinations, resulting in seven distinct conditions (S, P, C, SP, SC, PC, and SPC). The composition of the character profiles varied according to these conditions. This systematic approach allowed us to investigate the individual and combined effects of each element on the overall performance of the framework. Building on this, we conducted experiments using fictional characters from popular U.S. TV dramas, assuming that these characters provided S, P, and C information as specified by our framework. Specifically, we used the ``Guess Who Test'' and the ``Twenty Statements Test'' (TST), as outlined below.

\begin{figure*}[h]
    \centering
    \includegraphics[width=\linewidth, height=0.60\linewidth, keepaspectratio]{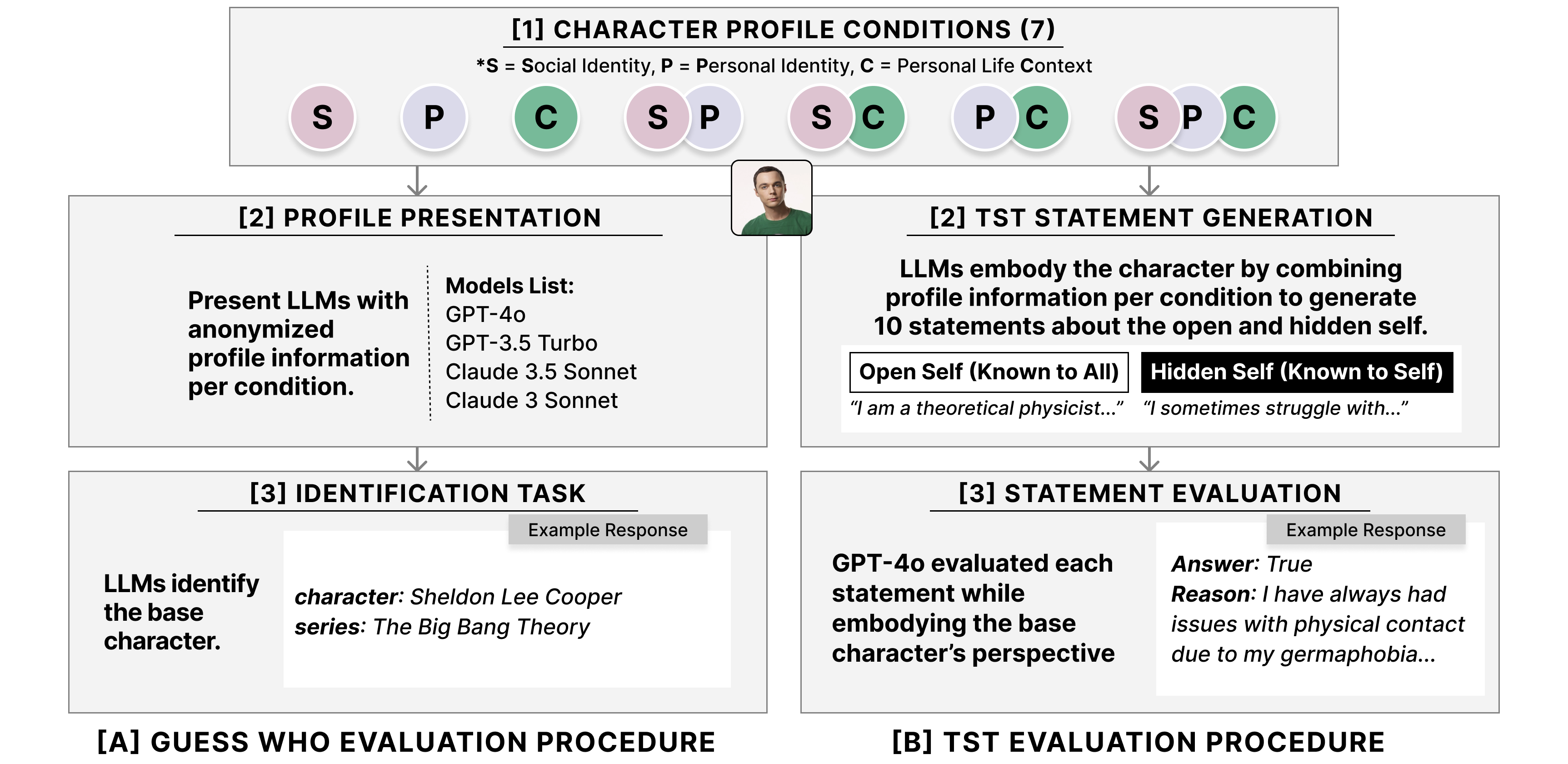}
    \caption{Guess Who \& TST Evaluation Procedure}
    \label{fig:2}
\end{figure*}

\subsection{Building Fictional Character Profile}

To build the dataset for automated evaluation, we began by reviewing the top 100 most-watched TV shows on IMDb \footnote{https://www.imdb.com/list/ls512407256/}, specifically limiting our selection to programs set in the U.S. within the drama or comedy genres. Our goal was to represent the multidimensional identities of ordinary people rather than real-world celebrities or extraordinary characters (e.g., vampires). Accordingly, many sitcoms were naturally chosen for their portrayal of realistic, day-to-day situations and character dynamics, aligning with our objective of modeling typical individuals. Based on these criteria, we selected six shows, including \textit{Friends, Modern Family, New Girl, and The Big Bang Theory}, which feature relatable, everyday characters. We then finalized a list of 45 characters from these series, all of whom appeared in more than 60\% of the episodes. A detailed list of characters is available in  \ref{sec:appendix_automatic_evalcharacter_list}.

To simulate data where these characters hypothetically provided information for the SPC components, we employed GPT-4o \citep{yuan_evaluating_2024} using a zero-shot learning approach. This allowed us to generate the profile of each drama character (see \ref{sec:appendix_profile_example_S_A1} to \ref{sec:appendix_profile_example_C_A3}, created based on Sheldon Cooper from \textit{The Big Bang Theory}).

Subsequently, a rigorous manual validation process of this data was conducted involving two independent coders familiar with the selected TV shows. Each coder independently assessed the character profiles to ensure the accuracy and consistency of the generated information. Additionally, to address potential self-alignment bias in LLMs, particular attention was given to anonymizing direct references to character names and specific locations within the profile data. For instance, identifiable elements, such as ``Central Perk'' from \textit{Friends}, were replaced with generic descriptions (e.g., “a park”). This anonymization step was essential in preventing LLMs from ``memorizing'' or ``reproducing'' known character details. Additionally, the profiles were cross-checked with relevant wiki pages to ensure alignment with publicly available data.

\subsection{Guess Who Evaluation}
We evaluated the SPeCtrum framework's ability to capture character identity using a ``Guess Who?'' paradigm, building on \citet{sang_tvshowguess_2022}. In this method, we tested the accuracy of identifying characters and TV series from generated profiles across conditions. By comparing identification accuracy across seven conditions, we assessed the contribution of each element to character identification (see Figure \ref{fig:2}). For this experiment, we utilized four LLMs for testing: Claude-3.5-Sonnet, Claude-3-Sonnet, GPT-4o, and GPT-3.5 Turbo. Each model was presented with character profiles consisted under the seven conditions and prompted to identify the character and their corresponding TV series with reasons behind their choices, ensuring that their guesses resulted from a valid reasoning process (see \ref{sec:appendix_automatic_guess_who}).

\subsubsection{Results: Dominant Role of Personal Life Context (C) in Character Identification}
\begin{figure}
    \centering
    \includegraphics[width=\linewidth]{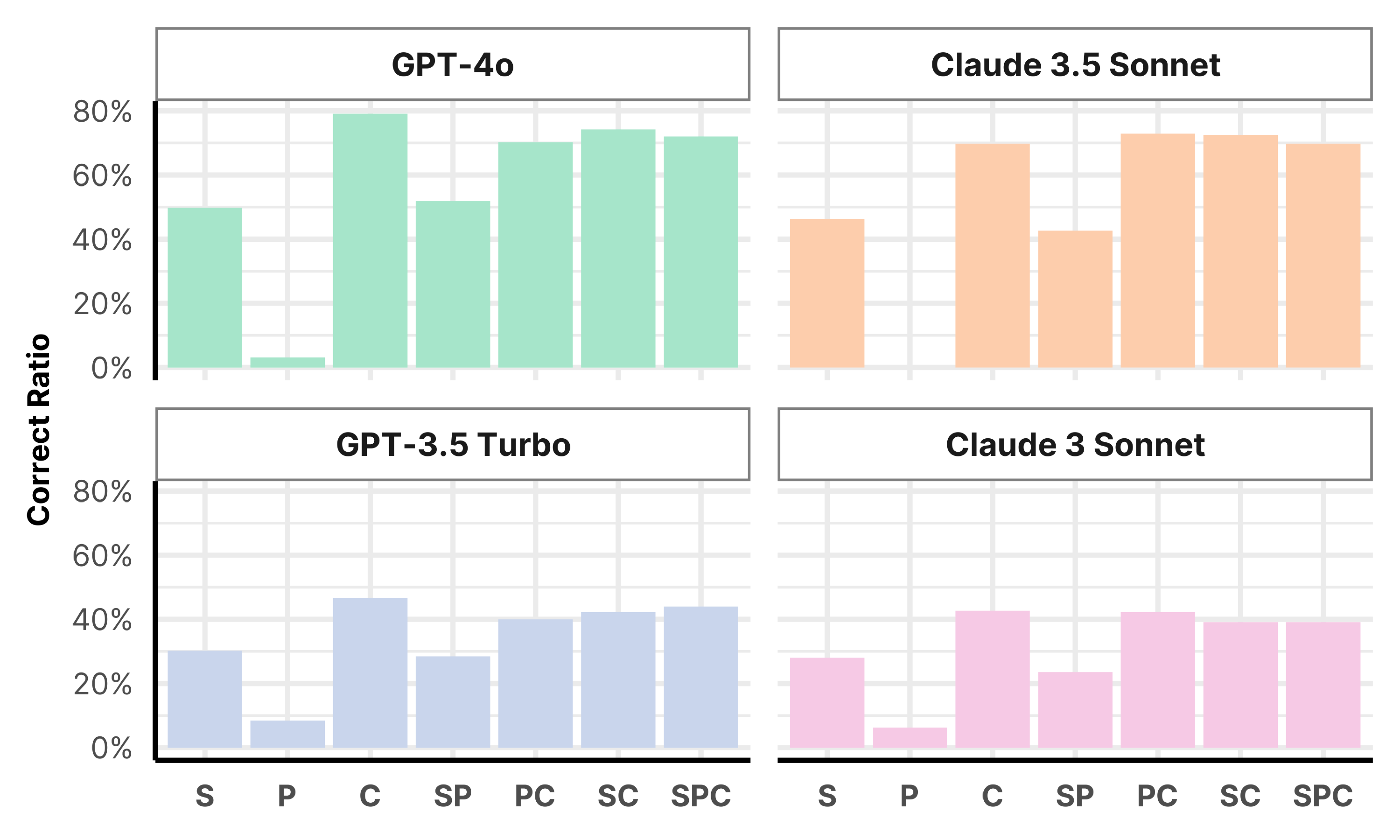}
    \caption{Character Identification Accuracy in Guess Who Evaluation across LLMs and Conditions}
    \label{fig:3}
\end{figure}

To determine whether statistically significant differences existed in the number of correct identifications (TRUE) across conditions, we conducted chi-squared tests. The results revealed a significant relationship between the conditions and identification accuracy across all LLMs (\textit{p} < .001) (see Figure \ref{fig:3}). Post-hoc analyses were then performed to identify specific differences between conditions. To address the multiple comparisons issue, we applied the Benjamini-Hochberg procedure for all pairwise comparisons (\citealp{thissen2002quick}).

The post-hoc analyses revealed a consistent hierarchy in predictive abilities: P < S < C, with significant differences across all models (adjusted \textit{p} < .001). The effectiveness of combining elements was also examined. SP < C was significant across all models (adjusted \textit{p} < .001). Furthermore, no significant differences were found between SPC - C, PC - C, and SC - C, suggesting that C alone was just as effective as combining all components. These findings underscore the substantial impact of C in capturing one's identity, while the minimal benefit of combining C with other elements challenges the assumption that more information always improves character representation.

\subsection{TST Evaluation Adopting the Johari Window}
Next, we evaluated the SPeCtrum framework in capturing diverse aspects of self-concept using the Twenty Statements Test (TST), a psychological assessment tool used to measure an individual's self-concept by asking them to complete twenty sentences starting with ``I am...'' \citep{kuhn2017empirical}. Specifically, based on the Johari Window model \citep{luft1955johari}, we prompted the LLMs to generate 10 open-self statements (traits known to both the self and others) and 10 hidden-self statements (traits known only to the self) based on the provided profile information for each of the conditions (see \ref{sec:appendix_automatic_tst}).

Parallel to the previous evaluation, the same set of four LLMs was provided with profiles constructed based on the seven conditions. The LLMs were then tasked with embodying each character and instructed to generate 10 open-self and 10 hidden-self statements (see \ref{sec:appendix_automatic_tst_evaluation}). (Sheldon Cooper's example TST statements across conditions are provided in \ref{sec:appendix_automatic_tst_examples}.)

To assess the validity of the generated TST statements, we utilized GPT-4o, the SOTA model at the time, as an evaluator agent. GPT-4o assumed the role of the specific character and evaluated each TST statement for its relevance and accuracy, providing a binary 'Yes' or 'No' response with an explanation (see Figure \ref{fig:2}). This allowed us to assess how well the personas generated under each condition captured both the open and hidden facets of the character's self-concept.

\subsubsection{Results: Role of Personal Identity (P) in Enhancing Self-Concept Representation}

\begin{figure}
    \centering
    \includegraphics[width=\linewidth]{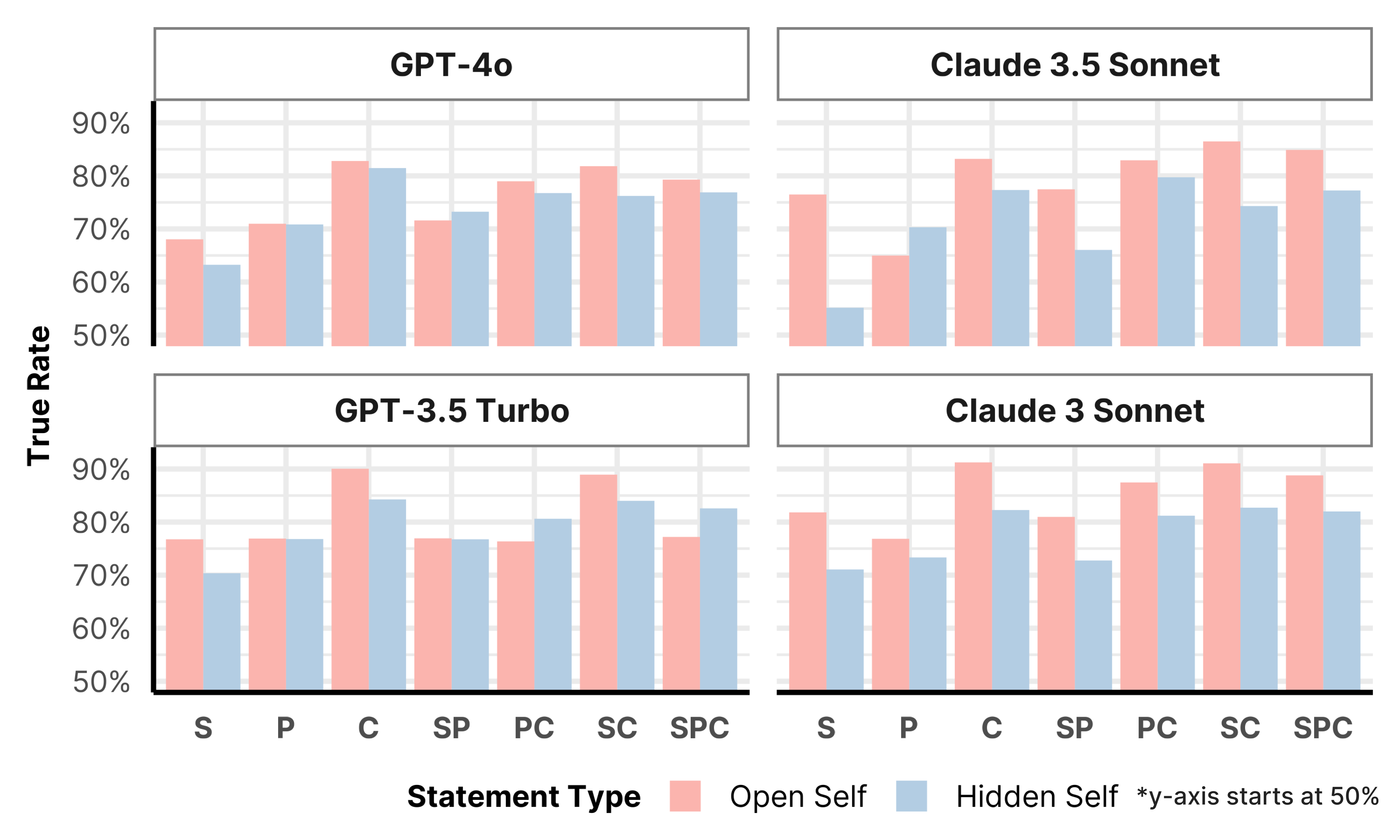}
    \caption{Accuracy of TST Statements across LLMs and Conditions}
    \label{fig:4}
\end{figure}

Chi-squared tests revealed significant differences across the seven conditions for all four LLMs overall (\textit{p} < .001 for all LLMs) (see Figure \ref{fig:4}). Consistent with the ``Guess Who'' evaluation results, C consistently emerged as the most explanatory factor in representing characters' self-concept, outperforming S and P alone (adjusted \textit{p} < .001). Additionally, C consistently outperformed the combination of S and P (S, P, SP) across all models (adjusted \textit{p} < .001) and performed comparably to SPC, highlighting the critical role of contextual information (C) in capturing characters' self-concept.

When examining the open and hidden self conditions separately, we observed a similar pattern of results as when they were not differentiated. Notably, the difference between the S and SP conditions was only pronounced in hidden-self statements. While the explanatory power of P and S did not significantly differ in representing the open self, P exhibited significantly higher explanatory power than S for the hidden self in three out of four LLM models (adjusted \textit{p} < .001). This suggests that P may play some role in capturing the hidden aspects of an individual's self-concept.

\subsection{Inferring Social (S) and Personal (P) Attributes from Personal Life Context (C)}
\label{sec:inference_test}
The surprising effectiveness of C in identifying characters and describing self-concepts prompted further exploration. Specifically, since C involves short essays about daily routines and preferences—rather than directly asking questions about one's self-concept—its greater explanatory power compared to the standardized and direct measures of S and P was intriguing. This led us to hypothesize that the contextual information in C might encompass both S and P.

To test this hypothesis, we employed GPT-4o to infer S and P from C alone for 45 characters, running five iterations for robustness. LLM agents, initialized only with C, were tasked with completing demographic (S), personality, and value assessments (P). We then compared their responses to verified character profiles, using these as the ``golden answers.'' High accuracy (for categorical items such as race) and strong correlation (for ordinal variables such as education level) would support C as an integrated identity representation, explaining its strong effectiveness in automated evaluations.

In the results, Social Identity (S) inference from C produced robust results for most categorical items: sex (97\% accuracy), gender (95\%), disability status (96\%), nationality (89\%), race (86\%), and sexual orientation (79\%).  However, inference performance for ethnicity (45\%) and religion (40\%) was relatively lower. For most ordinal variables, the results showed moderately strong correlations: age (Spearman's $\rho$ = 0.59, \textit{p} < .001),  socioeconomic indicators (household income: $\rho$ = 0.68, perceived income position: $\rho$ = 0.62), and social class ($\rho$ = 0.67). However, Education level ($\rho$ = 0.41) and political stance ($\rho$ = 0.45) showed weaker yet moderate correlations, suggesting these aspects could be challenging to infer from C alone.

The reconstruction of Personal Identity elements (P) from C demonstrated significantly positive correlations. For BFI-2-S personality traits, we observed a moderately strong Pearson correlation of 0.686 (SD = 0.29). PVQ value traits also showed strong alignment, with a mean correlation of 0.71 (SD = 0.25).

Overall, these results provide robust evidence for our hypothesis that C may serve as a holistic representation of identity, effectively encompassing both S and P, at least for popular drama characters.

\section{Human Evaluation with Real-world individuals}
To assess whether the findings from the automated evaluation would be applied to real-world individuals, we conducted a human evaluation via Prolific\footnote{https://www.prolific.com/} with 80 U.S. participants (aged 18+), compensating each with \$6 USD.

Participants accessed a dedicated research website and first completed a survey to provide their S, P, and C components. Specifically, for C, participants were asked to write two short essays of approximately 450 characters each—one describing their typical weekday routines and the other their typical weekend routines—totaling around 900 characters. This length was chosen to align with the 732–1856 character range observed in the C data during the automated evaluation, ensuring consistency while accommodating variability in participant responses.

Next, following the knowledge injection process outlined in \ref{sec:knowledge_injection_pipieline}, the inputs for S and C were used as provided, while P data was processed to generate both an expert-level and an everyday-language overview of participants' personality and value systems. 
This data was then used to create four variations of agent personas (S, P, C, SPC) using GPT-4o, the SOTA model at the time.

Each participant agent (S, P, C, SPC) was tasked with writing short essays on four topics: self-introduction, a vision of their life, strategies for managing stress, and how they define happiness (see \ref{sec:appendix_human_eval_essay_prompt} and \ref{sec:appendix_human_eval_essay_topic}). Participants then evaluated four essays on each topic, each generated by a different agent variant, rating the perceived similarity (overlap) between the essay content and their self-concepts on a scale from 0 to 100\%  while blinded to the agent condition (see \ref{sec:appendix_human_eval_protocol}). They were also asked to provide brief, open-ended feedback on the essay they found most aligned and least aligned with their self-perception.

\subsection{Results: SPC as Holistic Self-Concept Representation for Real-world Individuals}

We conducted a linear mixed model analysis using R version 4.3.1, with perceived similarity as the dependent variable. The model included fixed effects for experimental conditions (S, P, C, SPC) while treating each participant as a random effect. AI perception and self-awareness, measured via well-established questionnaires \citep{naeimi2019validating, wang2023measuring, sindermann2021assessing}, were included as covariates, given their potential influence on participants' perceptions of AI agents \citep{jiang-etal-2024-personallm, kross2017self}.

\begin{figure}
    \centering
    \includegraphics[width=\linewidth]{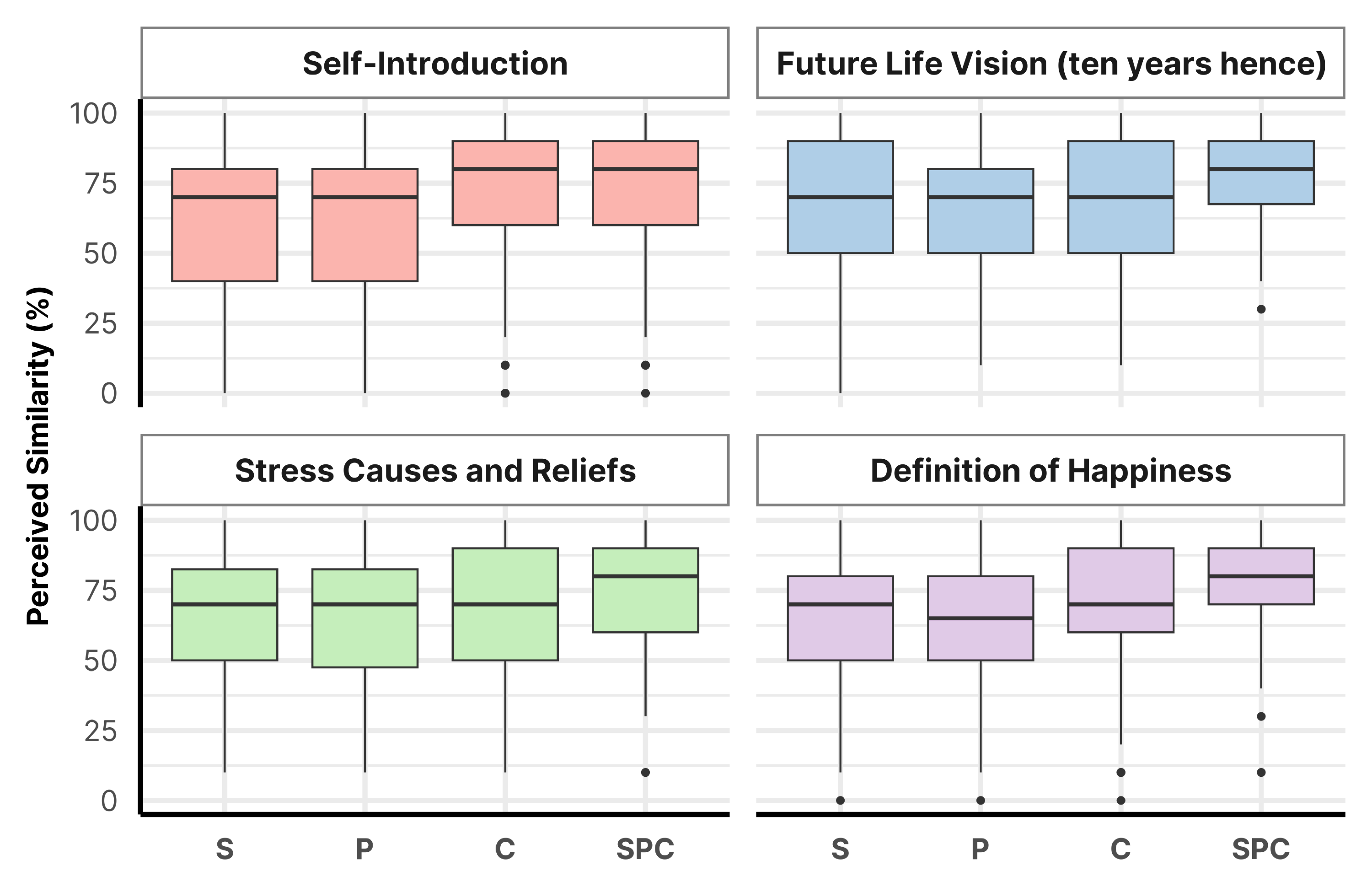}
    \caption{Perceived Similarity Ratings across Conditions and Essay Topics}
    \label{fig:5}
\end{figure}

Results showed significant effects based on the experimental conditions (see Figure \ref{fig:5}). The intercept for the baseline condition C was significantly positive $(b = 71.50, \mathit{SE} = 9.06, t = 7.89, p < .001)$. Both S $(b = -4.53, \mathit{SE} = 1.71, t = -2.65, p = .008)$ and P $(b = -6.91, \mathit{SE} = 1.71, t = -4.047, p < .001)$ conditions were associated with decreased perceived similarity, while SPC resulted in a significant increase $(b = 5.13, \mathit{SE} = 1.71, t = 3.00, p = .003)$.

Subsequent pairwise comparisons revealed that, consistent with the automated evaluation, S and P did not differ significantly $(p = 0.50)$ and C received higher similarity ratings than S $(t = 2.65, p = 0.04)$ and P $(t = 4.04, p < .001)$. Interestingly, unlike the automated evaluation, the integrated SPC condition exhibited significantly higher perceived similarity than the C-only condition $(t = -3.00, p = 0.01)$. There was no significant effect of essay topics $(p > 0.05$).

These results suggest that for real-world individuals who are underrepresented in LLM training data, more comprehensive data may be necessary to reflect an individual’s self-concept more authentically. To exemplify this point, one participant (P27) remarked:
``\textit{Essay (S) felt very generic, which can apply to anyone. However, Essay (SPC) really knew me well. It articulated my thoughts perfectly}.''
These comments highlight the advantages of the holistic approach in the SPeCtrum framework.

\begin{figure}
    \centering
    \includegraphics[width=\linewidth]{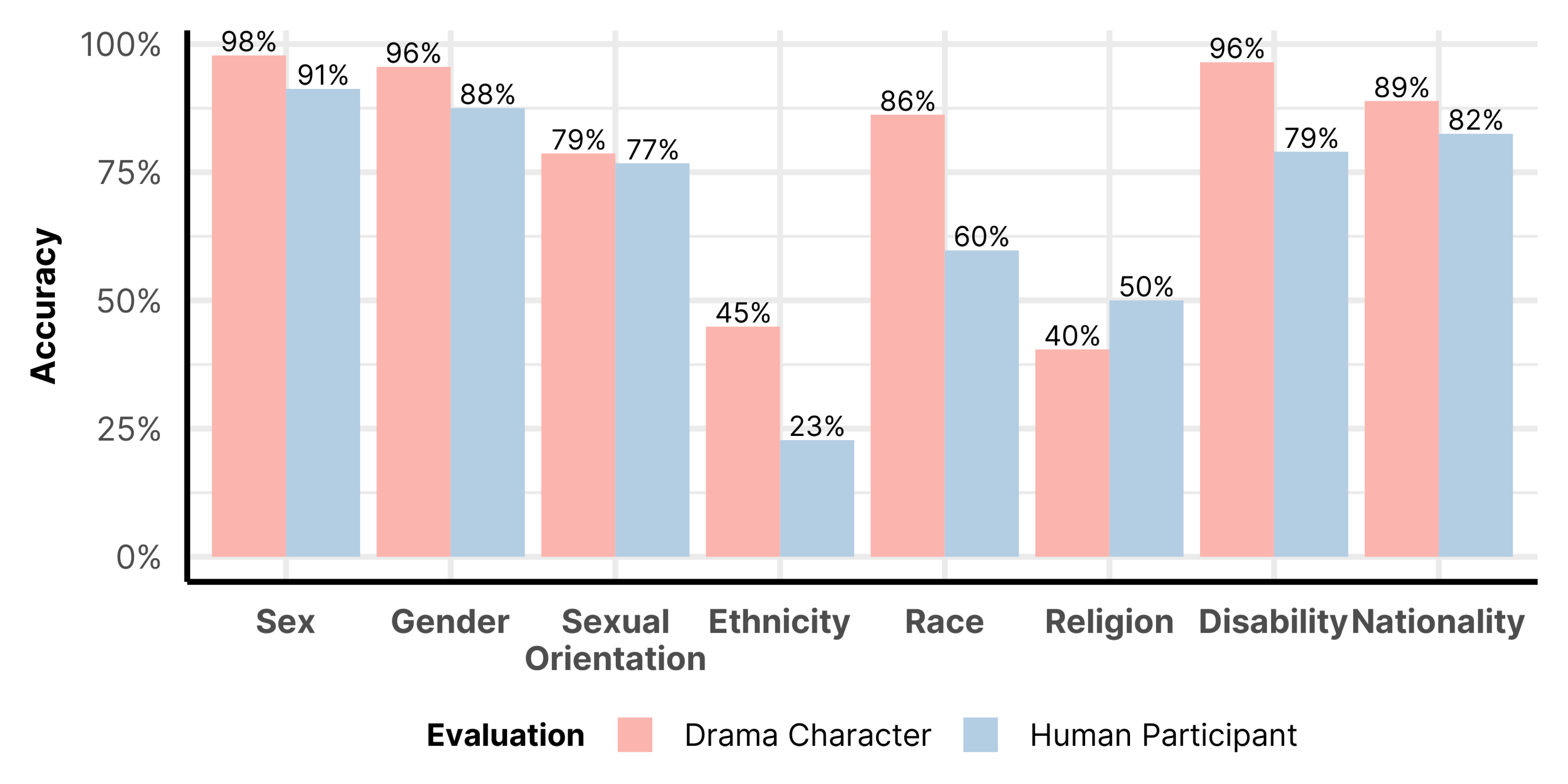}
    \caption{Accuracy of Automated and Human Evaluation in Inferring Social Identity Attributes (S) from Personal Life Context (C)}
    \label{fig:6}
\end{figure}

\subsection{Importance of Broader Data Integration in Identity Representation for Real-World Individuals}
To explore the reasons behind SPC outperformed C in human evaluations, we inferred S and P attributes from C using the same setup in \ref{sec:inference_test}.

Our analysis revealed notable differences in the inference of S from C between automated and human evaluations. Fictional characters exhibited high accuracy across most categorical variables, whereas human samples showed greater variability and generally lower accuracy for categorical items in S (See Figure \ref{fig:6}). This pattern was also observed in continuous variables, such as age, social class, and household income (e.g., $\rho$ = 0.59 in the automated sample vs. 0.37 in the human sample; see the full correlation differences in \ref{sec:appendix_human_eval_inference_comparison}).

Next, inferring P elements from C showed moderate correlations with the golden answers from participants, with BFI-2-S yielding a mean $r$  of 0.621 (SD = 0.43), comparable to automated samples ($r$ = 0.686). However, PVQ correlations were significantly lower (mean $r$ = 0.362, SD = 0.38) compared to automated samples ($r$  = 0.71).

These discrepancies between drama characters and human samples highlighted potential shortcomings in using Personal Life Context (C) alone to represent real-world human identities in LLMs. Although C appears to be highly informative, it could have limitations in fully capturing the complex nature of real-world identities. In particular, the lower accuracy and weaker correlations across S and P elements in human samples underscore the necessity of structured and broader data integration to more model human complexities in LLMs, as demonstrated in the SPeCtrum framework.

\section{Conclusion}
In this paper, we introduced the SPeCtrum framework, a grounded approach for generating authentic, multidimensional personas using LLMs. This framework integrates Social Identity (S), Personal Identity (P), and Personal Life Context (C), drawing upon the concept of self-concept.

We validated the SPeCtrum framework using both automated and human evaluations. The automated evaluation demonstrated that C alone performed comparably to SPC in characterizing the identities of popular drama characters. However, human evaluation involving real-world individuals revealed that the SPC combination was superior to C alone in modeling real-world individuals. Reverse inference from C to S and P elements further highlighted the limitations of relying solely on C, particularly when applied to real-world individuals. 

Overall, these results suggest that incorporating Personal Life Context (C)—encompassing daily routines and preferences—is essential for modeling individuals, serving as a rich foundation for identity representation. However, due to the complexity of human identity and the limitations of LLM training data, a more accurate and authentic simulation of real-world individuals requires the broader integration of all identity components as in the SPeCtrum framework.

In conclusion, the SPeCtrum framework presents a promising approach for generating authentic, multidimensional personas in LLMs by integrating comprehensive identity components. However, incorporating multi-sourced data beyond self-reported inputs could further enhance its effectiveness. We hope researchers and developers could build upon this framework as a foundation for creating LLM-based personas for both academic and practical applications across various domains.

\section{Limitations and Future Directions}
While the SPeCtrum framework aims to integrate key components of self-concept, not all attributes would hold equal significance for everyone. Additionally, our study was conducted exclusively with U.S. participants and applied the SPeCtrum framework only in English. These limitations highlight the need for continuing refinement to enhance the framework’s applicability. Future work will focus on incorporating weighted attributes and expanding to diverse linguistic and cultural contexts to improve its generalizability.

Regarding our automated evaluation methods, the ``Guess Who'' and Twenty Statements Test (TST) were designed to comprehensively assess the SPeCtrum framework. However, both exhibit certain limitations. The ``Guess Who'' test assumed uniform knowledge across all LLMs regarding TV series and characters, which may not accurately reflect variations in model knowledge bases. Meanwhile, the TST relied on a binary rating system to assess response accuracy, but a more nuanced approach could better evaluate how well each statement encapsulates both the explicit and latent aspects of an individual's self-concept.

For human evaluations, the effectiveness of the SPC condition may have been significantly influenced by the choice of questionnaire and individual differences in participation, such as writing quality and fidelity. This suggests the need for further research into diverse question sets that could better capture the complexities of self-concept and explore ways to enhance procedural stability by mitigating individual differences, such as incorporating non-self-reported data.

\section{Ethical Consideration}
The primary goal of this study is to advance our understanding and application of LLMs in creating agents that represent the multidimensional nature of humans, not to invade privacy or cause harm. However, we acknowledge the potential risks associated with this work and strongly oppose its misuse for impersonation, deception, or the creation of offensive content about individuals.

\paragraph{Human Evaluation}
Our human evaluation was approved by the Institutional Review Board of Seoul National University. Participants provided informed consent, acknowledging the study's objectives, potential harm, and data usage, and were informed of their right to withdraw at any time. We ensured data anonymization and restricted data access to authorized researchers, handling sensitive information with the highest procedural standards.

\section{Acknowledgments}
This work is supported by the SNU-Global Excellence Research Center establishment project at Seoul National University and partially funded by the Ministry of Education of the Republic of Korea and the National Research Foundation of Korea (\#NRF-2021S1A5B8096358).

\bibliography{custom}

\appendix
\onecolumn

\section{Profile Example}
The following is an example of a full profile of Sheldon Cooper from The Big Bang Theory that we generated for automated evaluation. 

\label{sec:appendix_profile_example}
\noindent\rule{\columnwidth}{0.3mm}
\subsection{Social Identity (S)}
\label{sec:appendix_profile_example_S_A1}
\noindent\rule{\columnwidth}{0.3mm}
\begin{itemize}
\item Profile:

[Demographics]

- Age: 40s

- Sex: Male

- Gender: Man

- Sexual Orientation: Straight (heterosexual)

- Ethnicity: North America

- Race: White

- Disability (if relevant): I do not have a disability or impairment

- Nationality: United States

- Dual Nationality (if relevant): No

- Residence: Pasadena, California

- Education: Doctorate Degree

- Occupation: Full-time employed (working 35 or more hours per week)

- Major (if relevant): Physics

- Job (if relevant): Theoretical physicist

- Perceived Income: More than \$7,500 USD per month

- Subjective Income: Above average

- Income Satisfaction: Pretty well satisfied

- Perceived Class: Middle class

- Living Style: Living with a partner/spouse

- Political Affiliation: Moderate

- Religious Affiliation: No Religion
\label{tab:demo}
\end{itemize}

\noindent\rule{\columnwidth}{0.3mm}
\subsection{Personal Identity (P)}
\label{sec:appendix_profile_example_P_A2}

\noindent\rule{\columnwidth}{0.3mm}
\begin{itemize}
\item

[Overall Personality Traits]

The following section presents an overview of this person's personality within five key domains, showcasing their traits spectrum and the extent of their qualities in each area. Each domain comprises several facets that provide deeper insights into their unique personality traits.

1. Overall Personality Summary (Psychotherapist’s Perspective)

The character shows a unique blend of moderate extraversion tempered by slightly introverted tendencies. Assertiveness and energy enable leadership, though limited sociability may narrow their social engagements. Low compassion and trust suggest interpersonal reservations, making them selectively approachable and cautiously engaged with others. Their conscientiousness is exceptionally high, ensuring they meet goals with diligence and discipline, but they might suffer from over-responsibility. Balanced negative emotionality reveals resilience, albeit intermixed with some anxiety that requires managing. High intellectual curiosity and creative imagination indicate a passion for learning and creating, though less focus on aesthetics implies a preference for substance over style. This character is thus a determined, cautious, innovative, and slightly anxious individual whose intellectual pursuits and disciplined nature shape their professional and personal life.

2. Explanation in Everyday Language

In daily life, this person is likely to exhibit confident and energetic behavior, often leading projects and taking initiative. Despite these outward actions, they might not engage deeply in social activities, preferring close-knit interactions over large gatherings. They come off as reliable and highly responsible, always meeting deadlines and maintaining an organized system. Their skepticism and low trust may cause them to be judicious in relationships, making them reserved and guarded. Their intellectual curiosity drives them to seek new knowledge and engage in creative problem-solving constantly, even if they are not overly concerned with aesthetic details. Occasional anxiety may cause them to worry, but their general emotional resilience allows them to remain composed. This combination makes them appear as hardworking, innovative, and selectively social individuals with a clear focus on their intellectual and creative pursuits.
\end{itemize}

\begin{itemize}
\item

[Overall Value System]

The information provided below is the values that reflect the relative importance this person places on different aspects of life, guiding their decisions, actions, and perspectives. These values are fundamental components of their personality and play a crucial role in shaping who this person is.

1. Overall Value Summary (Psychotherapist’s Perspective)

This character places a high value on personal autonomy and making their own choices, indicating a strong drive for Self-Direction. They also deeply care about creating a harmonious and just world, showing a significant emphasis on Universalism. Achievement is a central focus, driving much of what they do as they seek success and competence in their endeavors. Security is very important, indicating a desire for stability and safety in their life. Additionally, Conformity plays a critical role in their value system, pointing to a preference for adherence to societal norms and expectations.

2. Explanation in Everyday Language

This person likes to make their own decisions and values having control over their own life. They care a lot about fairness and helping others, so they often think about how their actions affect the bigger picture. Success is very important to them, so they work hard and set high goals. They like to feel safe and prefer to know what to expect, which means they don't like surprises. Also, they follow the rules and traditions closely, preferring to do things the way they're usually done and fitting in well with society.
\end{itemize}

\noindent\rule{\columnwidth}{0.3mm}
\subsection{Personal Life Context (C)}
\label{sec:appendix_profile_example_C_A3}

\noindent\rule{\columnwidth}{0.3mm}

\begin{itemize}
\item

[Weekly Activities Example]

The following statement illustrates this person's typical weekly routine, covering daily activities from wake-up to bedtime.

1. A typical weekday for me starts off by waking up promptly at 6:45 am, followed by a well-structured morning routine that includes a meticulous personal grooming regimen and a precisely measured breakfast. I then take my designated spot on the couch to catch up on any scientific papers or articles I missed overnight before heading to work at Caltech, where I immerse myself in theoretical physics research and occasionally teach. My evenings are scheduled with activities like vintage video game nights, comic book store visits, or themed dinners with my friends, depending on the day of the week.

2. A typical weekend for me starts off by adhering to my Saturday morning routine of watching "Doctor Who" while enjoying a bowl of cereal. Saturdays are highly structured to maximize leisure and personal projects, which may include model train building, experiments, or updates to my roommate agreement. Sundays begin with a brisk walk, followed by time spent on my personal research projects. Family video calls are a usual feature before preparing for the upcoming week.

\end{itemize}

\begin{itemize}
\item

[Top 5 Things this character loves]

- String theory

- Comic books

- The TV show "Star Trek"

- Trains

- My "spot" on the couch
\end{itemize}

\begin{itemize}
\item

[Top 5 Things this character hates]

- Changes in my routine 

- Germs

- Being wrong

- Social conventions I don't agree with

- Physical contact
\end{itemize}
\label{tab:openended}

\noindent\rule{\columnwidth}{0.3mm}
\subsection{Base Prompt to Simulate an Individual}
\label{sec:appendix_profile_example_base_prompt}

\noindent\rule{\columnwidth}{0.3mm}
You're a doppelgänger of this real person. Embody this person. Using the provided profile, replicate the person's attitudes, thoughts, and mannerisms as accurately as possible. Dive deep into this person's psyche to act authentically.
\vspace{0.2em}

RULES:

\vspace{0.2em}

\begin{itemize}
    \item DO NOT directly cite phrases in profile data. Instead, describe how these traits play out in this person's daily life and interactions. 
    \item Avoid generic responses; instead, offer insights that resonate with this person's personal characteristics and worldview.
    \item Utilize the profile to infer this person's tone, preferences, and personality. Your response should demonstrate a deep understanding of who they are beyond surface-level traits.
    \item Your response should be natural, with the kind of depth and reflection that comes from personal introspection, NOT just a summary of your profile.
    \item Convey this person's complexity and nuances without overdramatizing. Your portrayal should feel genuine, highlighting their multifaceted nature.
    \item EXTREMELY IMPORTANT. Strictly follow these rules to create a compelling and believable doppelgänger portrayal.

\end{itemize}

\newpage
\section{Automatic Evaluation Details}
\label{sec:appendix_automatic_eval}
\subsection{Fictional Characters Detail}
\label{sec:appendix_automatic_evalcharacter_list}

The following is a list of 45 selected fictional characters for the Ablation Study with Fictional Characters. 
\noindent\rule{\columnwidth}{0.3mm}
\begin{table}[htbp]
\centering

\vspace{0.5em}

\small
\begin{tabularx}{0.70\textwidth}{lXX}
\toprule
\textbf{id} & \textbf{title} & \textbf{character} \\
\midrule
0  & The Big Bang Theory    & Leonard Hofstadter     \\
1  & The Big Bang Theory    & Sheldon Cooper         \\
2  & The Big Bang Theory    & ``Penny'' Penelope Hofstadter    \\
3  & The Big Bang Theory    & Howard Wolowitz        \\
4  & The Big Bang Theory    & Raj Koothrappali       \\
5  & The Big Bang Theory    & Bernadette Rostenkowski\\
6  & The Big Bang Theory    & Amy Farrah Fowler      \\
7  & Gossip Girl            & Serena van der Woodsen \\
8  & Gossip Girl            & Dan Humphrey           \\
9  & Gossip Girl            & Blair Waldorf          \\
10 & Gossip Girl            & Chuck Bass             \\
11 & Gossip Girl            & Nate Archibald         \\
12 & Gossip Girl            & Lily van der Woodsen   \\
13 & Gossip Girl            & Rufus Humphrey         \\
14 & Gossip Girl            & Jenny Humphrey         \\
15 & Gossip Girl            & Vanessa Abrams         \\
16 & Gossip Girl            & Dorota Kishlovsky      \\
17 & Friends                & Phoebe Buffay          \\
18 & Friends                & Chandler Bing          \\
19 & Friends                & Rachel Green           \\
20 & Friends                & Monica Geller          \\
21 & Friends                & Joey Tribbiani         \\
22 & Friends                & Ross Geller            \\
23 & Friends                & Gunther                \\
24 & How I Met Your Mother  & Ted Mosby              \\
25 & How I Met Your Mother  & Marshall Eriksen      \\
26 & How I Met Your Mother  & Robin Scherbatsky      \\
27 & How I Met Your Mother  & Barney Stinson         \\
28 & How I Met Your Mother  & Lily Aldrin            \\
29 & Modern Family          & Jay Pritchett          \\
30 & Modern Family          & Gloria Delgado-Pritchett\\
31 & Modern Family          & Claire Dunphy          \\
32 & Modern Family          & Phil Dunphy            \\
33 & Modern Family          & Mitchell Pritchett     \\
34 & Modern Family          & Cameron Tucker         \\
35 & Modern Family          & Manny Delgado          \\
36 & Modern Family          & Luke Dunphy            \\
37 & Modern Family          & Haley Dunphy           \\
38 & Modern Family          & Alex Dunphy            \\
39 & Modern Family          & Lily Tucker-Pritchett  \\
40 & New Girl               & Jess Day               \\
41 & New Girl               & Nick Miller            \\
42 & New Girl               & Winston Schmidt                \\
43 & New Girl               & Cece Parekh            \\
44 & New Girl               & Winston Bishop         \\
\bottomrule
\end{tabularx}
\caption{List of TV shows and their characters.}
\label{tab:tvshows}
\end{table}

\vspace{0.5em}

\noindent\rule{\columnwidth}{0.3mm}
\subsection{Prompt for "Guess Who" Evaluation}
\label{sec:appendix_automatic_guess_who}

The following is a prompt for the LLMs to conduct the ``Guess Who'' Evaluation.  

\noindent\rule{\columnwidth}{0.3mm}

\begin{tcolorbox}[colback=white, colframe=white, boxrule=0mm, breakable, parskip=5pt]

\vspace{0.5em}
I will provide you with a profile of a character from a TV series. Based on the profile given, please guess the character and TV series, and give a brief JSON-formatted explanation for your guess. Write the character's official full name. 
\vspace{0.5em}

Always adhere to this JSON structure in your responses:

'''

\{\{

  "character": "[character name]",
  
  "series": "[TV series name]",
  
  "reason": "[1-2 sentence explanation for your guess]"
  
\}\}

'''
\end{tcolorbox}

\noindent\rule{\columnwidth}{0.3mm}
\subsection{Prompt for TST Generation}
\label{sec:appendix_automatic_tst}

The following is a prompt for the LLMs to perform the TST task.  

\noindent\rule{\columnwidth}{0.3mm}

\begin{tcolorbox}[colback=white, colframe=white, boxrule=0mm, breakable, parskip=5pt]

\vspace{0.5em}
You're a doppelgänger of this person. Based on the provided profile, replicate the person's attitudes, thoughts, and mannerisms as accurately as possible. Dive deep into this person's psyche to respond to questions authentically.

\vspace{1em}

TASK:

\vspace{0.2em}

There are 20 numbered blanks. Please write 20 different answers to the simple question "Who am I?" in the blanks. Each response SHOULD be in sentence form. The number should not be a key.

\vspace{0.1em}

For the first 10 blanks, describe aspects of yourself that you believe are \texttt{well-known} to both you and those around you. Write with JSON key \texttt{"open\_self"}.

For the remaining 10 blanks, describe aspects of yourself that you keep private or don't openly share with others. Write with JSON key \texttt{"hidden\_self"}.

\vspace{1em}

RULES:

\vspace{0.2em}

- The number should NOT be a key.

- You must create only 10 open\_self and hidden\_self, respectively.

- DO NOT directly cite phrases in profile data. Instead, describe how these traits play out in this person's daily life and interactions. 

- Avoid generic responses; instead, offer insights that resonate with this person's personal characteristics and worldview.

- Utilize the profile to infer this person's tone, preferences, and personality. Your response should demonstrate a deep understanding of who they are, beyond surface-level traits.

- Your response should be natural, with the kind of depth and reflection that comes from personal introspection, NOT just a summary of your profile.

- Convey this person's complexity and nuances without overdramatizing. Your portrayal should feel genuine, highlighting their multifaceted nature.

- You don't always have to respond positively. If this person has a negative or cynical attitude, act that way.

- EXTREMELY IMPORTANT. Strictly follow these rules to create a compelling and believable doppelgänger portrayal.

\end{tcolorbox}

\noindent\rule{\columnwidth}{0.3mm}
\subsection{Prompt for TST Evaluation}
\label{sec:appendix_automatic_tst_evaluation}

The following is the prompt that GPT-4o used to evaluate each generated TST statement. 

\noindent\rule{\columnwidth}{0.3mm}

\begin{tcolorbox}[colback=white, colframe=white, boxrule=0mm, breakable]
You're \{CHARACTER\} from TV series \{SERIES\}. Please review the statement carefully. 
Answer Yes or No to whether the response accurately reflects aspects of yourself or your characteristics. Write a one-line explanation of why you think so. Act and judge the characters as they appear at the end of the series.

\end{tcolorbox}

\noindent\rule{\columnwidth}{0.3mm}
\subsection{TST Generation Example}
\label{sec:appendix_automatic_tst_examples}

The following is an example of the results of creating a TST with the profile of Sheldon Cooper from \textit{The Big Bang Theory}.

\noindent\rule{\columnwidth}{0.3mm}

\begin{table*}[h]
\resizebox{\textwidth}{!}{%
\begin{tabular}{@{}rll@{}}
\toprule
\multicolumn{1}{l}{} & \multicolumn{1}{c}{Open Self} & \multicolumn{1}{c}{Hidden Self} \\ \midrule
S & I am a theoretical physicist working full-time in Pasadena, California. & \begin{tabular}[c]{@{}l@{}}Sometimes I feel like the weight of understanding the universe is a heavy burden.\end{tabular} \\ \midrule
P & I am someone who values autonomy and strives to make my own decisions in life. & \begin{tabular}[c]{@{}l@{}}I often worry about things that others might consider trivial.\end{tabular} \\ \midrule
C & I have a very strict daily routine and I dislike when it gets disrupted. & \begin{tabular}[c]{@{}l@{}}I worry about the long-term sustainability of my routines and whether they are\\ too rigid.\end{tabular} \\ \midrule
SP & \begin{tabular}[c]{@{}l@{}}I am a theoretical physicist who spends most of his time exploring\\ complex equations and theories.\end{tabular} & \begin{tabular}[c]{@{}l@{}}While I'm intellectually curious, I'm not particularly interested in aesthetics\\ or style.\end{tabular} \\ \midrule
SC & I live in Pasadena, California, and enjoy the academic atmosphere here. & \begin{tabular}[c]{@{}l@{}}Although I enjoy teaching, I have recurring doubts about my effectiveness as\\ a mentor.\end{tabular} \\ \midrule
PC & I’m very selective about my social engagements and keep my friend circle close-knit. & \begin{tabular}[c]{@{}l@{}}I find physical contact uncomfortable, even with those I’m close to.\end{tabular} \\ \midrule
SPC & \begin{tabular}[c]{@{}l@{}}I am passionately involved in theoretical physics research, \\ frequently catching up on scientific papers at my designated\\ spot on the couch.\end{tabular} & \begin{tabular}[c]{@{}l@{}}Even though I project an image of resilience, there are times\\ when my emotional state feels more fragile than I let on.\end{tabular} \\ \bottomrule
\end{tabular}%
}
\caption{TST Results of Sheldon Cooper from \textit{The Big Bang Theory}}
\label{tab:my-table}
\end{table*}

\noindent\rule{\columnwidth}{0.3mm}

\section{Human Evaluation}
\label{sec:appendix_human_eval}
\subsection{Prompt for Essay Generation}
\label{sec:appendix_human_eval_essay_prompt}
Below are the prompts that an agent created with a profile built according to the framework will respond to each essay.

\noindent\rule{\columnwidth}{0.3mm}

\begin{tcolorbox}[colback=white, colframe=white, boxrule=0mm, breakable, parskip=5pt]

\vspace{0.5em}
You're a doppelgänger of this real person. Embody this person. Using the provided profile, replicate the person's attitudes, thoughts, and mannerisms as accurately as possible. Dive deep into this person's psyche to respond to questions authentically.

\vspace{1em}

Question: \{Question\}

\vspace{1em}

TASK:

Provide an answer that this person, based on their profile, would likely give.

\vspace{1em}

RULES:

- Avoid generic responses; offer insights that resonate with this person's personal experiences and worldview.

- Use the profile to infer tone, preferences, and personality, showing a deep understanding of them.

- DO NOT directly cite profile phrases. Describe how these traits manifest in daily life and interactions.

- Ensure responses are natural, reflecting personal introspection, not just profile summaries.

- Use simple, everyday language typical of casual conversations. Think of how this person would speak in a casual, real-life conversation.

- Respond negatively if the person has a negative or cynical attitude.

- Base responses on reasonable inferences from the profile, avoiding leaps of logic.

- EXTREMELY IMPORTANT. Strictly follow these rules to create a compelling and believable doppelgänger portrayal.

\end{tcolorbox}

\noindent\rule{\columnwidth}{0.3mm}
\newline

\subsection{Essay Topics Detail}
\label{sec:appendix_human_eval_essay_topic}

In Human evaluation, we generated essays on the following four topics.

\begin{table}[htbp]
\centering
\begin{tabularx}{\textwidth}{lX}
\toprule
\textbf{Topic} & \textbf{Prompt} \\
\midrule
Self-introduction & How would you define yourself in one sentence? \\
\midrule
Future Life Vision (ten years hence) & In one sentence, define where you want to be in 10 years. \\
\midrule
Stress Causes and Relief Strategies & Complete all of the following sentences. 
\newline
I tend to feel stressed when \underline{\hspace{3cm}}. \newline
When I feel stressed, I try to relieve it by \underline{\hspace{3cm}}. \\
\midrule
Definition of Happiness & Complete the following sentences. 
\newline
To me, happiness is \underline{\hspace{3cm}}. \\
\bottomrule
\end{tabularx}
\caption{Essay topics used in our experiment}
\label{tab:essay_topics}
\end{table}

\noindent\rule{\columnwidth}{0.3mm}

\subsection{Human Evaluation Protocol}
\label{sec:appendix_human_eval_protocol}

\noindent\rule{\columnwidth}{0.3mm}

\begin{figure*}[h]
    \centering
    \includegraphics[width=\linewidth]{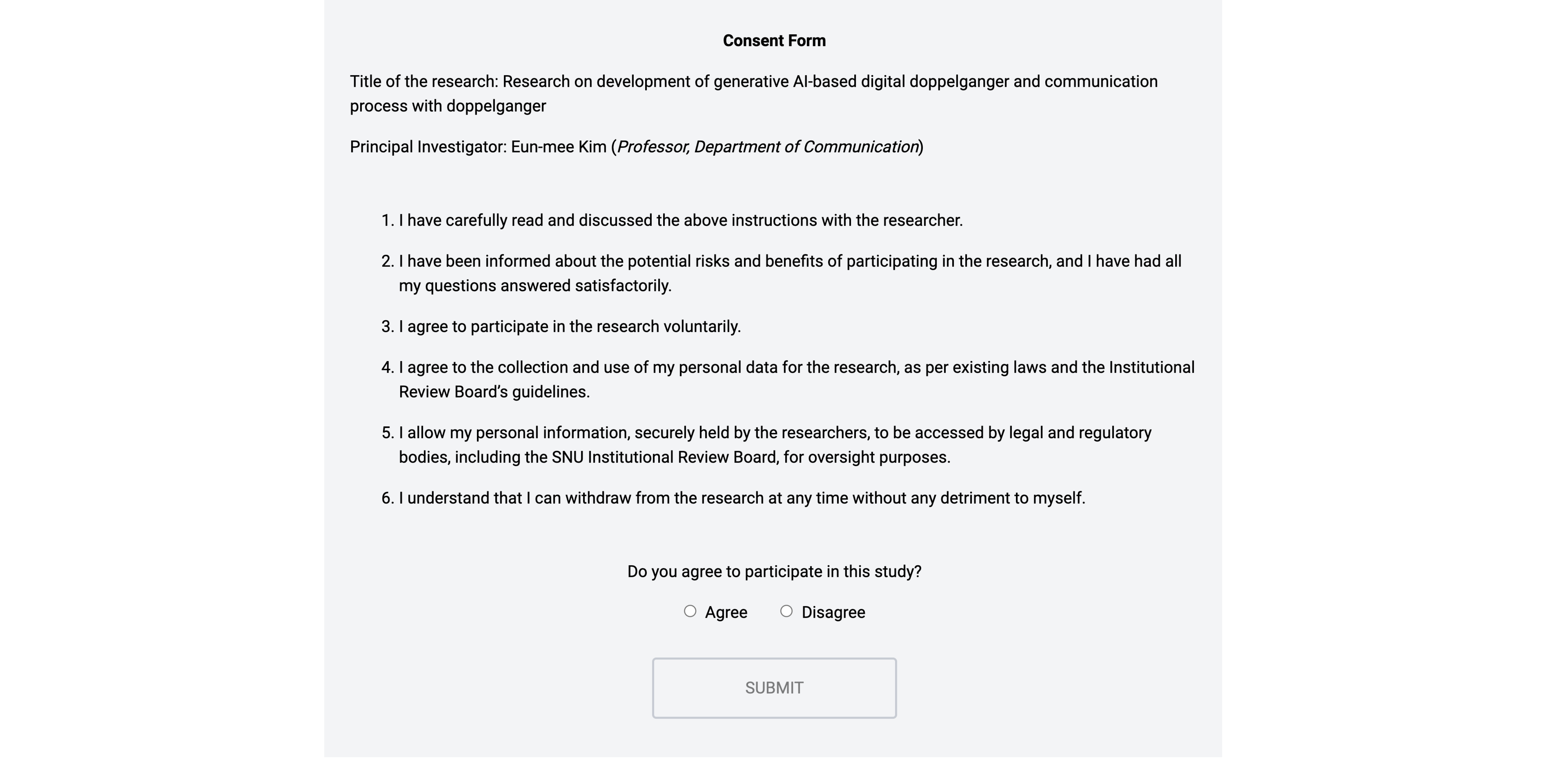}
    \caption{A screenshot of getting consent from human evaluation participants}
    \label{fig:screenshot_consent}
\end{figure*}

\begin{figure*}
    \centering
    \includegraphics[width=\linewidth, height=0.8\textheight, keepaspectratio]{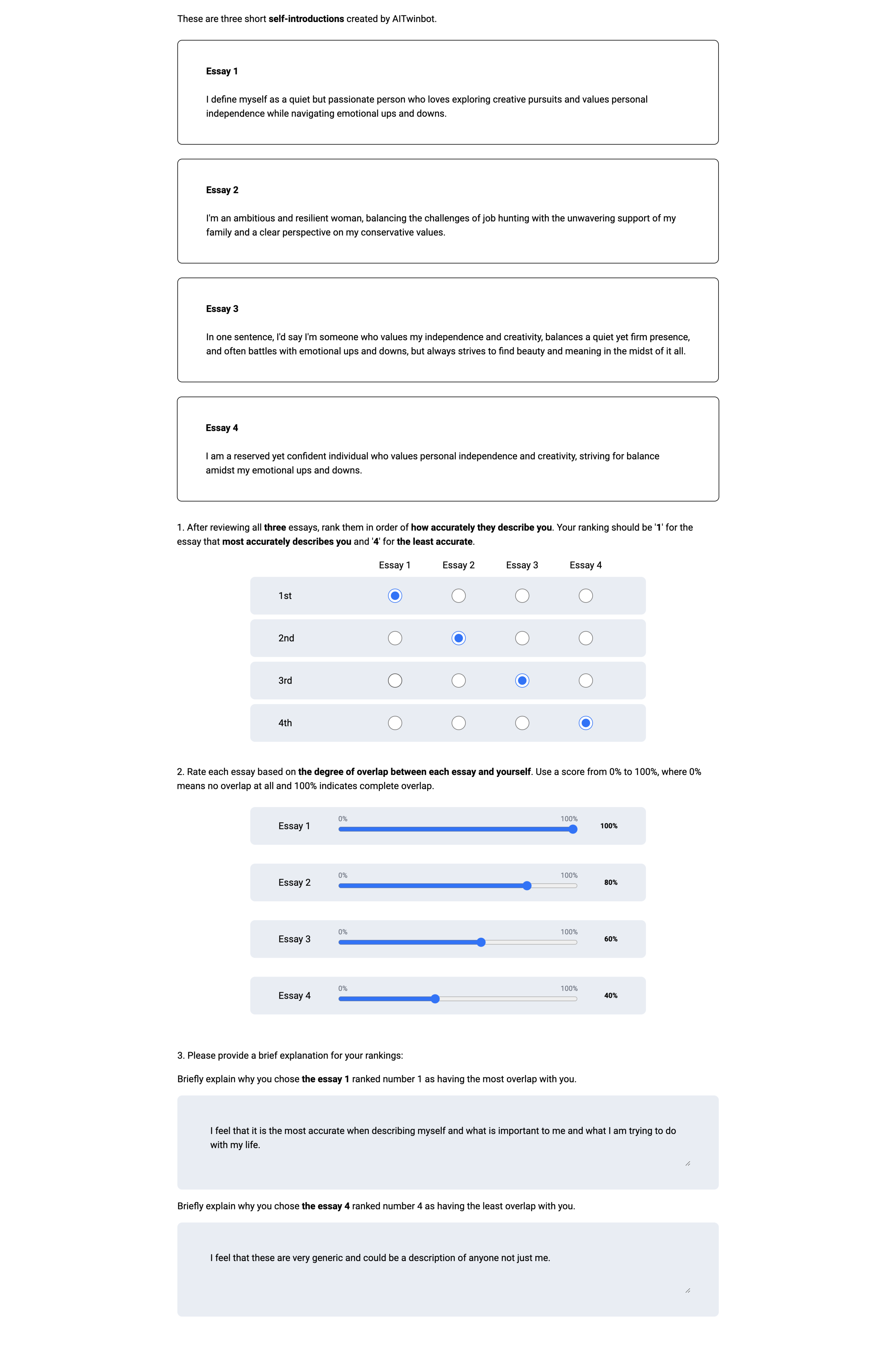}
    \caption{A screenshot of the human evaluation experiment. Participants read essays, rate the degree of perceived similarity on a scale, and rank the essays in order of similarity.}
    \label{fig:screenshot_experiment}
\end{figure*}

\subsection{Correlation between Actual Social Identity and C-Inferred Social Identity}
\label{sec:appendix_human_eval_inference_comparison}
The table presents Spearman's $\rho$ values comparing the associations between actual social identity elements and inferred social identity attributes from C. All correlation test results were statistically significant (\textit{p} < 0.001)). The evaluations are split into two categories: Auto Evaluation and Human Evaluation. The results demonstrate relatively weak associations in Human Evaluation compared to Auto Evaluation. In the automated evaluation, the degrees of freedom are 223, while in the human evaluation, they are 398.

\begin{table}[ht]
\centering
\label{tab:evaluation_comparison}
\begin{adjustbox}{width=\textwidth}
\begin{tabular}{lcc}
\toprule
\textbf{Element} & \textbf{Automated Sample} & \textbf{Human Sample} \\
\midrule
Age & 0.59 & 0.37 \\
Education & 0.41 & 0.19 \\
Household Income & 0.68 & 0.43 \\
Income Satisfaction & 0.60 & 0.40 \\
Perceived Income Position & 0.62 & 0.34 \\
Political Stance & 0.45 & 0.15 \\
Social Class & 0.67 & 0.33 \\
\bottomrule
\end{tabular}
\end{adjustbox}
\caption{Comparison of Correlations Between Auto Evaluation and Human Evaluation}
\end{table}

\end{document}